\documentclass{article}

\PassOptionsToPackage{hyphens}{url}
\PassOptionsToPackage{numbers}{natbib}
\usepackage[preprint]{neurips_2020}
\usepackage[bookmarks=false]{hyperref}
\usepackage{cleveref}
\usepackage{longtable}
\usepackage{geometry}
\usepackage{graphicx}
\usepackage{qtree}
\usepackage{booktabs}
\usepackage{siunitx}
\usepackage{ragged2e}
\usepackage{rotating}
\usepackage{adjustbox}
\usepackage{caption}
\usepackage[T1]{fontenc}
\usepackage{pdflscape}
\newcolumntype{R}[1]{>{\RaggedRight}p{#1}}
\usepackage{amssymb}
\usepackage[table]{xcolor}

\title{Formatting Instructions For NeurIPS 2020}

\author{%
  Nina Effenberger\thanks{corresponding author} \\
  Cluster of Excellence Machine Learning\\
  University of Tübingen\\
  72076 Tübingen \\
  \texttt{nina.effenberger@uni-tuebingen.de} \\
  \And
  Nicole Ludwig \\
  Cluster of Excellence Machine Learning\\
  University of Tübingen\\
  72076 Tübingen \\
  \texttt{nicole.ludwig@uni-tuebingen.de}
}

\begin{document}

\title{A Collection and Categorization of Open-Source Wind and Wind Power Datasets}
\maketitle

\begin{abstract}
Wind power and other forms of renewable energy sources play an ever more important role in the energy supply of today's power grids. Forecasting renewable energy sources has therefore become essential in balancing the power grid. While a lot of focus is placed on new forecasting methods, little attention is given on how to compare, reproduce and transfer the methods to other use cases and data. One reason for this lack of attention is the limited availability of open-source datasets, as many currently used datasets are non-disclosed and make reproducibility of research impossible. This unavailability of open-source datasets is especially prevalent in commercially interesting fields such as wind power forecasting. However, with this paper we want to enable researchers to compare their methods on publicly available datasets by providing the, to our knowledge, largest up-to-date overview of existing open-source wind power datasets, and a categorization into different groups of datasets that can be used for wind power forecasting. We show that there are publicly available datasets sufficient for wind power forecasting tasks and discuss the different data groups properties to enable researchers to choose appropriate open-source datasets and compare their methods on them.  
\end{abstract}


\maketitle

\section{Introduction}

Over the past few years, the absolute and relative amount of sustainable and renewable energy sources integrated into the power grid has grown. Among these energy sources, wind power is the most common in many areas such as the European Union \cite{Renewableseu}. Wind power is both renewable and sustainable as it has only a minor impact on the environment, and its main resource, the wind, is not exhaustible. However, wind power is less reliable than unsustainable energy sources such as gas or coal due to the stochasticity of wind and weather in general. Furthermore, wind power cannot be generated when there is no wind, and alternative sources of energy have to be included into the power system to balance out the wind fluctuations. Therefore, sustainable energy forecasting is essential when it comes to predicting times at which alternative energy sources or measures to change the demand behaviour have to be taken in order to stabilize the grid.

Wind power forecasting is hence crucial for an efficient interplay between the different kinds of power and can be divided into different tasks. Among these tasks are predicting the actual power generation, variability of the wind or quick and large changes in the power generation \cite{giebel2017wind}. Independent of the forecasting task, wind power forecasting can be performed on different time scales, ranging from very short ($\leq 30$ minutes) to long-term (a day to a month) and on different spatial scales, ranging from individual wind turbines to whole regions \cite{hanifi2020critical}. While much literature focuses on these different wind power forecasting tasks on different spatial and temporal scales, little attention is given to the datasets used as the basis of this research. However, the underlying datasets are crucial, as using different datasets for the same tasks does not allow properly comparing the models and keeping the datasets private, instead of publicly available, renders reproducibility impossible.

In order to tackle this lack of comparability, Kariniotakis et al. \cite{kariniotakis2004performance} compare different models using different but fixed datasets. They reveal a dependence between the performance of the models and the complexity of the terrain on which the wind farm is situated. The authors show that depending on the turbine's environment, the normalized mean absolute errors (NMAEs) computed on the same time horizon can be more than three times higher (NMAE$>35\%$ in contrast to NMAE$<10\%$) for turbines on complex terrain in contrast to flat terrain. Additionally to the environment of a turbine, other close-by turbines influence the power output of individual wind turbines \cite{nygaard2014wakes}. These turbine-turbine interactions play a role in both, offshore and on-shore wind power forecasting (e.g. Barthelmie et al. \cite{barthelmie2010quantifying}, McKay et al. \cite{mckay2013wake}). Thus, different terrain complexities and turbine-turbine interactions make it hard or even impossible to compare methods tested and validated on different datasets. Additionally, prediction errors depend on the forecast horizon \cite{marti2006evaluation} and it is well known that the choice of data influences the result. However, we argue that these issues can be overcome by comparing models using the same open-source data.

Previous work has examined several of these open-source datasets. For example, Menezes et al. \cite{menezes2020wind} present datasets with a focus on wind farms and resources, most of the datasets they describe are open-source and some of them are also part of this survey. While the authors cover a broader range of wind resource datasets, we focus on wind power data and cover various additional and new datasets. Clifton et al. \cite{clifton2018wind} provide an overview of energy-related wind (and solar) resource datasets, covering onshore and offshore wind. Such wind resource data aims to quantify the amount of wind available for conversion into wind power and can, for example, be used to identify potential construction sites for future wind farms. However, Clifton et al. \cite{clifton2018wind} do not include any real turbine level datasets. 

The lack of literature giving an overview of currently available open-source datasets hinders open science. Confidential data limit the reproducibility and comparability of research and stand in contrast to the dogma of open research. In 2016, Wilkinson et al. \cite{wilkinson2016fair} proposed a set of principles and guidelines that should improve Findability, Accessibility, Interoperability, and Reusability of data. These so-called FAIR data principles are widely known, accepted and applied in various scientific research fields (e.g. El-Gebali et al. \cite{el2019pfam}, Sinaci et al. \cite{sinaci2020raw}, Vuong et al. \cite{vuong2018cultural}, Frank et al. \cite{frank2020fair}). Open-source data is not only seen as necessary in research; political institutions such as the Cabinet of Germany (Deutsche Bundesregierung) \cite{BRD} and the European Union \cite{EU} also emphasize the importance of open data. Nevertheless, these scientific and political ambitions are often not implemented, and the known issue of the result's dependence on the data is rarely addressed. 

Therefore, to bring forward open science in the wind power forecasting community, we present and categorize open-source datasets that can be used for wind power forecasting. Furthermore, our categorization and detailed descriptions simplify and motivate the use of non-confidential, open-source wind power data. The remainder of this paper is organized as follows: \Cref{sec:datasets} is the main focus of this paper and describes and categorizes the open-source datasets that contain wind and wind power data. We then evaluate the datasets properties in \Cref{areas} to help choose an appropriate dataset before discussing known systematic errors in wind data in \Cref{sec:syserrors}. We discuss our approach and selection of datasets in \Cref{discussion} and conclude in \Cref{conclusion}.

\section{Open-Source Wind (Power) Datasets}  \label{sec:datasets}

\begin{figure}[ht]
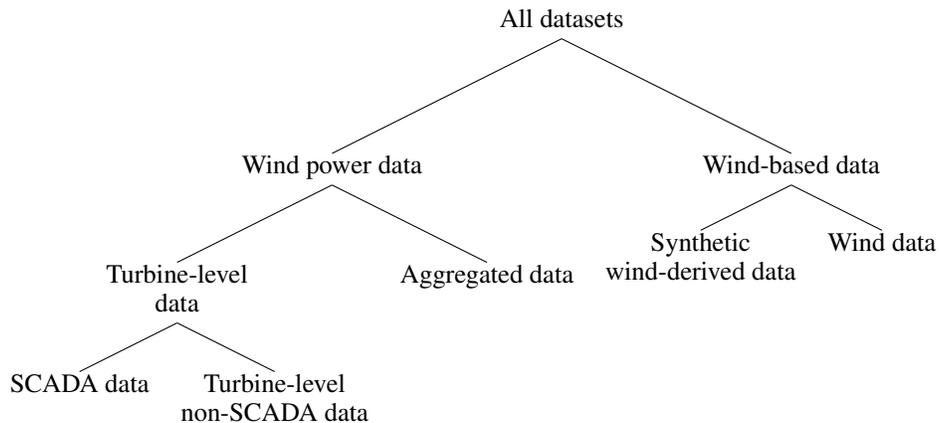
 \Tree [ .{All datasets} [ .{Wind power data} [ .{Turbine-level \\ data} [  .{SCADA data} ][ .{Turbine-level \\ non-SCADA data} ] ] [ .{Aggregated data} ] ] [ .{Wind-based data} [ .{Synthetic \\ wind-derived data} ] [ .{Wind data} ] ] ]
\caption{All of the presented datasets can be grouped into one of five different categories. The two super-groups are wind power data and wind-based data. The wind power datasets can be divided into the sub-groups turbine-level and aggregated data. Turbine-level datasets contain measurements on turbine level and aggregated data is spatially aggregated on different levels from farm to country level. All of the turbine-level datasets include wind and wind power measurements, SCADA data contains more variables than turbine-level non-SCADA data. The wind-based datasets contain either pure wind data or synthetic power data that is derived from wind data.
}
\label{fig:datagroups}
\end{figure}

In this section we give an overview of over forty datasets for wind power forecasting. We compiled the datasets listed in this paper in several different ways, mainly by searching online for datasets (e.g. on the webpages of the IEA 36 \cite{IEA36} or the WRAG\cite{wrag}), contacting researchers from different continents and looking into papers that work with disclosed data. We first introduce the groups into which we separate the datasets and give some insights into each group, before we present the groups separately. We mainly differentiate between two groups of data, namely \emph{wind power data} and \emph{wind-based data}. We then further split the wind power data into three sub-groups and the wind-based data into two sub-groups, resulting in a total of five different groups of data. Figure \ref{fig:datagroups} shows the connections between the five data groups. These five sub-groups have different characteristics, which allow us to assign each of the found open-source datasets into one of the groups. All data groups include wind data, but we can differentiate them using six different characteristics: whether they contain additional weather information, include real or synthetic wind power data, include turbine-specific control parameters, are on turbine level, and include turbine-level supervisory control and data acquisition (SCADA) information. \Cref{overviewgroups} gives an overview over the five data groups and their main characteristics. 

This distinction into five groups forms the basic framework for this paper and the rest of this section is therefore structured accordingly. We give some information on how our found datasets are distributed over the five data groups, the world and other descriptives in \Cref{ssec:stats}. We then introduce the datasets which we categorize as wind power data in \Cref{windpower}, before introducing the remaining datasets categorized as wind-based data in \Cref{windbased}. As we present the datasets in large tables we also introduce how these tables are structured and what information they contain in \Cref{ssec:guide}.

\small
\begin{table}[htb]
    \footnotesize
    \centering
    \captionsetup{font=footnotesize}
    \caption{Overview of the characteristics of the five different data groups. All of the presented datasets can be assigned to one of five data groups. The fine-grained datasets are SCADA data and other turbine-level data. Aggregated datasets do also contain wind power data but weather data is not included and can usually not be applied directly. The two last data groups are based on wind data and contain either synthetic power data or pure wind measurements.}
    \rowcolors{2}{gray!15}{white}
    \begin{tabular}{R{0.11\textwidth}R{0.08\textwidth}R{0.08\textwidth}R{0.08\textwidth}R{0.08\textwidth}R{0.08\textwidth}R{0.06\textwidth}R{0.06\textwidth}R{0.18\textwidth}} 
        \toprule 
         Data group & Wind data&Weather & Real wind power data & Synthetic wind power data&  Turbine-specific control parameters & Turbine-level data &SCADA data & Key characteristics \\ \midrule 
         SCADA data (Section \ref{windpower})& \checkmark& \checkmark&\checkmark&-&\checkmark&\checkmark&\checkmark&Includes SCADA data \\
         
         Other turbine-level data (Section \ref{windpower})& \checkmark&partly&\checkmark&-& rarely &\checkmark& - & Does not include SCADA data but turbine-level measurements \\
         
         Aggregated data (Section \ref{windpower})& \checkmark&partly &\checkmark&-&-&-&-&Data is aggregated spatially
         \\
         Meteorologically derived data (Section \ref{windbased})& \checkmark&\checkmark &-&\checkmark&-&-&-&No wind power measurements are contained \\
         
         Wind data (Section \ref{windbased})& \checkmark&\checkmark&-&-&-&-&-&No wind power data, only wind data \\
         \bottomrule
    \end{tabular}
    \label{overviewgroups}
\end{table}
\normalsize

\subsection{Descriptive Statistics} \label{ssec:stats}

\begin{figure}[ht]
    \centering
    \includegraphics[width=\textwidth]{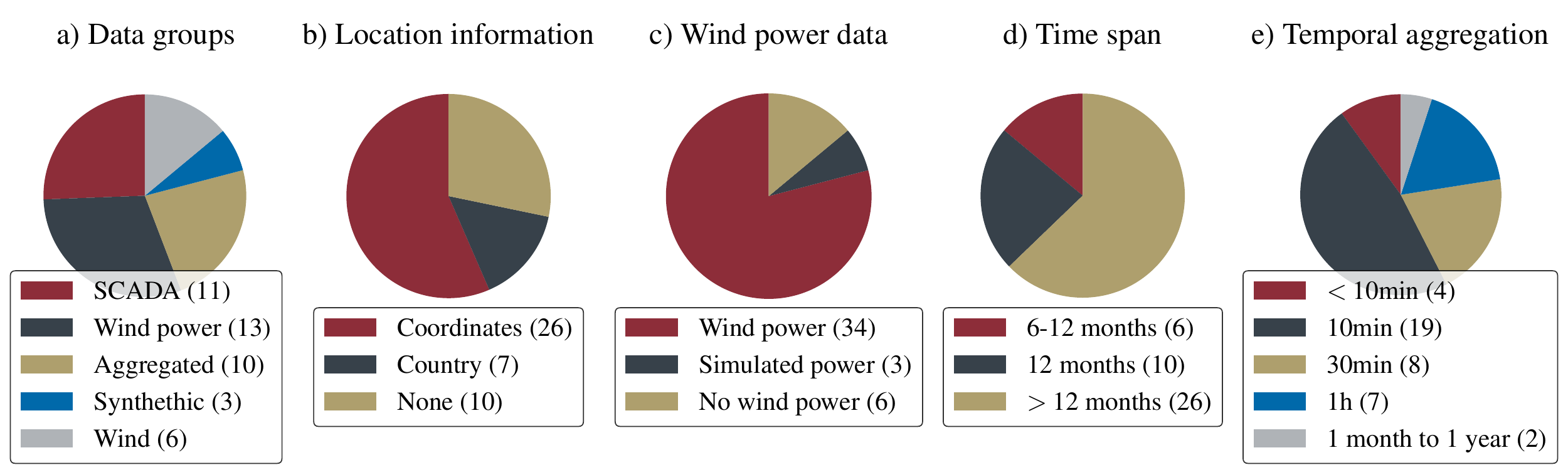}
    \footnotesize
    \caption{Descriptive statistics of the presented datasets. The numbers in the brackets display the absolute size of each group respectively. In a)-d) all of the five groups are contained, e) does not take the wind datasets into account. 
    }
    \label{fig:descriptive}
\end{figure}

The above described categorization into five groups is unique, meaning that each dataset is associated with exactly one group. In this subsection we want to take a look at the distribution of the datasets over these five groups and further properties that separate the datasets. \Cref{fig:descriptive} visualizes all properties and their composition within the entirety of the datasets. 

Regarding the data groups, most of the datasets contain SCADA (eleven datasets) or aggregated data (ten datasets), while the smallest group consists of synthetic data (three datasets). More than 75\% of the datasets contain real wind power data and more than 50\% of these data were collected on turbine-level. Location information is accessible for more than half of the datasets -- this allows for them to include additional weather data and most datasets cover at least one year of data. Vargas et al. \cite{vargas2019wind} investigate 145 different models and came to the conclusion that most models for long-term forecasting used hourly-data (49\%), 37 of the datasets that we present here provide or can be aggregated to this resolution. Most commonly in our selection of datasets, the data is provided in 10 minute intervals.

\begin{figure}[ht]
    \centering
    \includegraphics[width=\textwidth]{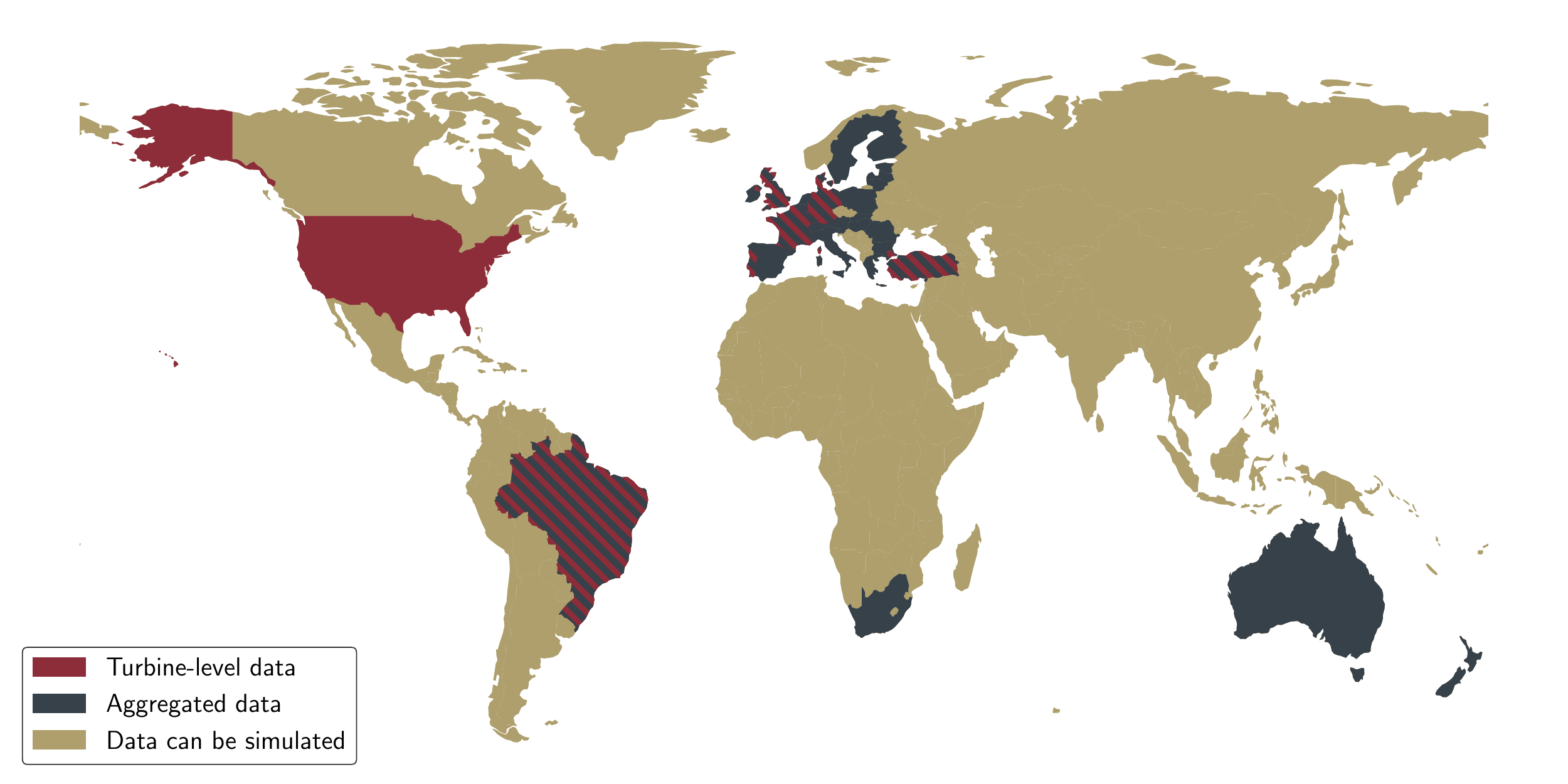}
    \caption{Country coverage of the datasets. Synthetic data can be simulated for all countries and locations given weather data. For several countries turbine-level data or aggregated data are accessible. Countries where at least one dataset of aggregated and turbine-level data are accessible are colored in stripes.}
    \label{fig:spatialcov}
\end{figure}

The datasets stem from different locations around the globe as presented on the map in \Cref{fig:spatialcov}. This map shows that there exist open-source datasets from all European countries, one achievement of the European association for the cooperation of transmission system operators for electricity (ENTSO-E) \cite{entsoe}. Furthermore the datasets also cover Africa, North and South America and Australia. However, among the datasets in this paper none are from Asia or the Poles. 

\subsection{Guide through the Tables} \label{ssec:guide}

\normalsize
We present all datasets in tables in their corresponding subsection. To make these tables easier to understand, we explain their structure in the following. The first column in each table, \textbf{Dataset}, gives an identifier of the dataset, which is, if possible, related to the location. We also provide a reference in this column, the first citation usually refers to the web page where the dataset is stored. If a paper exists, it is also referenced in this first column. Additional documentation or additional data can be found in the column \textbf{Information}. If accessible, the column \textbf{Location} provides the country were the data was collected or the region for which it was synthetically generated. The column \textbf{Coordinates} then gives the specific coordinates to this location, if available. The column \textbf{Time span} provides the time span the dataset covers and the column \textbf{Interval} the temporal distance between two successive data points. This interval is in general not equal to the distance between two measurements, but the measurements are averaged over this interval. Lastly, the column \textbf{Origin} gives an overview of the data providers and sources. Further, we provide the column \textbf{Additional information} which in general consists either of additional documentation, descriptions on how to add location data or variables that are included in the dataset. Additionally, if applicable, the number of turbines covered is given in the column \textbf{Turbines} and for the aggregated datasets the aggregation level is provided in the column \textbf{Aggregation level}. The column \textbf{Type} contains the type of wind data, which is either NWP model, reanalysis data or wind measurements. 

\subsection{Wind Power Data}
\label{windpower}
As stated above, we split the wind power data into three disjoint groups (see Figure \ref{fig:datagroups}): two of them contain turbine-level data that cover different variables, while the third group contains wind power data aggregated on different spatial scales.

In turbine-level data, weather variables such as wind direction are measured at hub height, the height on which wind power is generated. Having data at that height allows detecting turbulence patterns or effects of the surrounding area that data collected at lower or higher levels usually cannot detect. Additionally, wind power forecasts can also rely on other weather variables such as temperature or humidity \cite{lin2020wind}. Bilal et al. \cite{bilal2018wind} show that this additional weather information can increase the accuracy of the models. Finally, if the turbine location is known, the forecasting algorithm can include further weather data, such as numerical weather prediction (NWP) models. 

We further divide turbine-level data into two sub-groups. The first group includes supervisory control and data acquisition (SCADA) data. SCADA data covers a set of environmental, operational, thermal and electrical measurements \cite{astolfi2021multivariate} usually recorded for maintenance reasons. We give an overview of existing datasets that include SCADA data in \Cref{tab:SCADA}. The second data group covers non-SCADA turbine-level data, an overview of the datasets is given in \Cref{tab:nonSCADA}. The datasets in this non-SCADA group only account for a limited amount of weather variables and do not contain any technical measurements. All of them contain wind speed and wind power data and most of them also contain information on the wind direction.

However, among the three groups of wind power data the third group of datasets, namely aggregated datasets, are the ones that are most often publicly available. Aggregated data do not contain turbine-level measurements or turbine-specific power output but wind power data that is aggregated over different spatial regions, ranging from wind farms to whole countries. In contrast to turbine-level data, most of the datasets are of lower temporal resolution and contain hourly data. Nevertheless, the dataset in this category with the highest resolution is with measurements every 4 seconds the most finely resolved one in the whole collection. When the location of the wind farm is known, additional weather data can be taken into account. Several databases allow for a mapping of wind farms to their location. This mapping can for example be performed for the datasets provided by Elexon \cite{Elexon} and the transparency platform of ENTSO-E \cite{entsoe}, two of the largest aggregated datasets. Both are updated daily with a lag of approximately five days and wind power data can be found under "Actual Generation Output per Generation Unit". \Cref{tab:aggregated} gives more details of all aggregated datasets.

\subsection{Wind-based Data}
\label{windbased}
Wind power is generated by transforming the wind's kinetic energy into physical torque. As generated wind power is proportional to wind speed cubed \cite{grogg2005harvesting}, the performance of an operating wind turbine is mainly determined by wind speed. Some research therefore focuses on wind speed forecasting in order to perform wind power forecasting; sometimes this is also called indirect wind power forecasting \cite{zjavka2018direct}. \\
This dependence of wind power on wind speed is also exploited to generate meteorologically derived time series. These data are usually generated by transforming real wind and weather data into synthetic wind power data. By mapping wind intensity with a turbine-specific power curve to extracted wind power, data can be generated without taking any -- potentially disclosed -- wind power data of the turbines that are modeled into account. The wind datasets can be sub-divided into pure wind datasets and synthetic wind power datasets based on wind data. 
The datasets in \Cref{tab:wind} cover the former and contain either numerical weather predictions (NWPs), reanalysis data or met mast measurements. Reanalysis data consists of measured, post-processed and interpolated data \cite{rienecker2011merra} while NWPs use mathematical models of the environment and its current state to predict future weather variables.
Synthetic datasets that are based on wind data and can be derived from these can be found in Table \ref{tab:synthetic}. 
In general, all of the wind and wind power datasets are suitable for various wind power forecasting tasks, but not all of the datasets are suitable for every model. To find a suitable dataset for a specific model the variable description in the dataset tables (Table \ref{tab:SCADA}, \ref{tab:nonSCADA}, \ref{tab:aggregated}, \ref{tab:wind}, \ref{tab:synthetic}) should be considered.
\section{Evaluation of the Datasets' Properties} \label{areas}
Having introduced all datasets we take a closer look at the properties of the datasets and how they can be used in wind power forecasting. With this section we aim to enable choosing a proper data set for different wind power forecasting tasks. As the different data groups come with their group specific properties, we first address the tasks for which the datasets are often used and then discuss the shortcomings and advantages of each group. The discussion focuses on the role of specific variables as well as spatial and temporal granularity and coverage of the datasets. We do not evaluate the properties of the different wind measurements in this section but refer to \Cref{sec:syserrors} for this.
The goal of wind power forecasting is mainly threefold. It consists of wind farm site selection (1), efficiently harnessing the wind (2) -- mainly by controlling the turbine -- and efficiently integrating that generated wind power into the grid (3). Most other goals, such as maintenance planning, can be derived from results that tackle these three primary issues. To control the turbine and harness the wind efficiently (2), it is usually assumed that finely-resolved data on a second or even millisecond scale are needed \cite{giebel2017wind}. Therefore, even though eleven of the datasets contain control parameters, their temporal resolution of 10 minutes is not sufficient for efficient turbine control and we will not discuss turbine control in this review. 
Consequently, the tasks we cover aim to improve site selection (1) and wind power grid integration (3). The goal of the former is to optimize turbine or wind farm site selection under certain constraints. These constraints can be diverse; among them are the average wind speed, geological and geographical properties of the surrounding area, and political guidelines that, for example, restrict the closeness of turbines to cities \cite{rediske2021wind}. Furthermore, while site selection has to rely on pure wind and weather data only, wind power grid integration models usually also consider wind power data. 

\subsection{Wind Power Data}
Having introduced the two main goals that can be addressed using the datasets we present here, namely site selection and wind power grid integration, we will now first elaborate on the direct use of wind power data to perform wind power forecasting. The associated goal of wind power forecasting is usually efficient wind power grid integration. To achieve this goal, different sub-tasks can be defined. Among them are wind power time series forecasting on different temporal and spatial scales, ramp forecasting and variability forecasting. The main difference between these forecasting tasks is not always their underlying model or data but, in many cases, the evaluation metric used to assess the forecast quality. Therefore, the tasks mentioned here are neither disjoint nor do they necessarily require different models. An example of this can be found in Bianco et al. \cite{bianco2016wind}, where redefining the error metric allows using time series forecasts to investigate variability of the wind or ramp events. 

The underlying main task of wind power forecasting is usually wind power time series forecasting, a regression task that aims to predict wind power generation at future time points given historical data. Wind power time series forecasting is often divided into different types of forecasting, ranging from very short-term (in the range of a few minutes) to long term (in the range of one day to one month) \cite{hanifi2020critical}. Additionally, wind power forecasting models can also be classified by their prediction methodology (usually physical, statistical or hybrid) \cite{hanifi2020critical}.
Another sub-task of wind power forecasting is wind power ramp forecasting which aims to predict large and sharp variations in the wind and its associated wind power. While there is no unique definition of a wind ramp, such a ramp can be characterized by the direction and magnitude of the power variation and its corresponding duration \cite{ferreira2011survey}. Additionally, the rate at which such large variations occur and the time point at which they occur can play a role when evaluating wind ramps. Key characteristics of wind ramps are that they can lead to a power output that is far below or above the usual one which comes either with decreased power generation or potential damage of the turbine \cite{gallego2015review}. Both of these outcomes can have a negative impact on the energy supply.
The third task we mention here is wind variability forecasting. In contrast to wind power ramps, wind power variability refers to large amplitude, periodic changes in wind speed \cite{giebel2017wind}. However, there is no clear definition of wind variability either. Davy et al. \cite{davy2010statistical} introduce a variability index that "is defined as the standard deviation of a band-limited signal in a moving window". In general, wind fluctuations that are part of wind variability can for example exhibit climatic patterns with regard to the season or atmospheric processes such as cloud coverage. Having introduced these tasks, namely wind power time series forecasting, ramp forecasting and variability forecasting, we now elaborate the usefulness of the three groups of wind power data for these tasks. 
\begin{landscape}

\centering

\begin{table}
\footnotesize
\captionsetup{font=footnotesize}
\caption{Overview of the supervisory control and data acquisition (SCADA) datasets. The datasets contain SCADA data and the separation line divides the datasets with known location from the datasets without known location. A description of the variables can be found in the beginning of this Section. Additionally, the number of turbines that the dataset covers is provided in \textbf{Turbines}.}
\rowcolors{2}{gray!15}{white}
\begin{tabular}{R{0.1\textwidth}R{0.1\textwidth}R{0.1\textwidth}R{0.1\textwidth}R{0.1\textwidth}R{0.1\textwidth}R{0.35\textwidth}R{0.2\textwidth}}
\toprule
Dataset & Turbines & Location & Coordinates & Time span & Interval  & Additional information& Origin \\ \midrule    
\href{https://zenodo.org/record/1475197#.YS81CC-21hA}{Beberibe} \cite{Beberibe_PDS}& 32 & Brazil & \textit{Tabela} 20 of \textit{anexo} B \cite{sakagami2017influencia} & 08.2013-07.2014 & 10min & Used in Yoshiaki Sakagami's PhD thesis \cite{sakagami2017influencia} & Supported by the Brazilian Electricity Regulatory Agency \cite{Beberibe_PDS}\\ 
            
\href{https://zenodo.org/record/1475197#.YS81CC-21hA}{Pedra do Sal} \cite{Beberibe_PDS} & 20 & Brazil & \textit{Tabela} 19 of \textit{anexo} B \cite{sakagami2017influencia}& 08.2013-07.2014 & 10min & Used in Yoshiaki Sakagami's PhD thesis \cite{sakagami2017influencia} & Supported by the Brazilian Electricity Regulatory Agency \cite{Beberibe_PDS}\\

\href{https://zenodo.org/record/5946808}{Penmanshiel} \cite{plumley_charlie_2022_5946808} & 14 & United Kingdom & Can be found in the documentation \cite{plumley_charlie_2022_5946808}& 01.01.2016-01.07.2021& 10min & Additional data is provided on Zenodo \cite{plumley_charlie_2022_5946808} including site substation/PMU meter data and site fiscal/grid meter data. &  Cubico Sustainable Investments Ltd \cite{cubivo} \\

\href{https://data.dtu.dk/articles/dataset/Scada_data_from_Delabole_wind_farm/14077004}{Delabole} \cite{ncdel} & 10 & Great Britain  & Can be found in the documentation\cite{docdel}& 01.05.1993-30.04.1994& 10min & Documentation is available \cite{docdel}. & DTU Database \cite{DTU} \\ 

\href{https://zenodo.org/record/5841834}{Kelmarsh} \cite{plumley_charlie_2022_5841834} & 6 & United Kingdom & Can be found in the documentation \cite{plumley_charlie_2022_5841834}& 01.01.2016-01.07.2021& 10min & Additional data is provided on Zenodo \cite{plumley_charlie_2022_5841834} including site substation/PMU meter data and site fiscal/grid meter data. &  Cubico Sustainable Investments Ltd \cite{cubivo} \\

\href{https://opendata-renewables.engie.com/explore/index}{La Haute Borne} \cite{Hauteborne}& 4 & France & See static information \cite{Hauteborne}& 01.01.2013-13.01.2018 & 10min &  & Engie Renewables \cite{Hauteborneor} \\ 
                  
\href{https://data.dtu.dk/articles/dataset/Wind_resource_SCADA_data_and_time_series_of_wind_and_turbine_loads_from_Tjareborg_DK/16701961}{Tjareborg} \cite{tjaredata} & 1 & Denmark & 55.448233, 8.593803 \cite{doctjare} & 20.01.1988-18.01.1993 & 10min & Additional data of a close-by met mast is also available \cite{tjaremet}.  &DTU Database \cite{DTU} \\ 

\href{https://conservancy.umn.edu/handle/11299/205162}{Eolos Wind Research Station} \cite{eoloslink}\cite{davison2019}& 1 & United States &44.73, -93.05 \cite{davison2019evaluation}& 01.01.2017-31.12.2017& 10min & Incomplete SCADA data (74 different SCADA and other variables). See description on the same page for further details \cite{eoloslink}.& University of Minnesota (Brian Davison) \cite{davison2019}\\ 

\midrule

\href{https://opendata.edp.com/explore/dataset/wind-farm-1-signals-2016/information/?sort=-timestamp}{EDP} \cite{edp}& 4 & Portugal &unknown& 01.01.2016-31.12.2016& 10min &Registration necessary for data access. Description of the variables and additional data of a close-by met mast can be found on the same page. &EDP Inovação \cite{opendataedp} \\ 

\href{https://www.kaggle.com/theforcecoder/wind-power-forecasting}{Kaggle 1} \cite{wpf2} &  1 & Unknown & Unknown & 31.12.2017-31.03.2020 & 10min & & Unknown\\  

\href{https://www.kaggle.com/wasuratme96/iiot-data-of-wind-turbine}{Kaggle 2} \cite{Kaggle_FD} &  1 & Unknown & Unknown& 01.05.2014-09.04.2015 & 10min & Includes fault and status data & Unknown\\ 

\bottomrule
\end{tabular}
\label{tab:SCADA}
\end{table}

\end{landscape} 

\begin{landscape}

\centering

\begin{table}
\captionsetup{font=footnotesize}
\caption{Overview of non-SCADA turbine-level wind power datasets. The datasets contain wind power, wind speed and other turbine-level measurements but no full SCADA data. The separation line divides the datasets with known location from the datasets without known location. A description of the variables can be found in the beginning of this Section. Additionally, the number of turbines is provided in \textbf{Turbines}.}
\rowcolors{2}{gray!15}{white}
\footnotesize
\label{tab:nonSCADA}
\begin{tabular}{R{0.18\textwidth}R{0.07\textwidth}R{0.07\textwidth}R{0.16\textwidth}R{0.1\textwidth}R{0.1\textwidth}R{0.27\textwidth}R{0.2\textwidth}}
\toprule
Dataset & Turbines & Location & Coordinates & Time span & Interval  &Additional information & Origin \\ \midrule  \href{https://github.com/4castRenewables/climetlab-plugin-a6}{Maelstrom} \cite{maelstromdata}& 45 & Germany & Accessible after download & 6 months & 1h & Production, minimal
production, maximal
production,
mean wind speed,
minimal wind speed,
maximal wind speed max,
rotor speed,
minimal rotor speed,
maximal rotor speed,
errornumber,
Eisman regulation,
status\cite{maelstromdoc} & Notus Energy \cite{maelstromdata} \\ 
 
\href{https://www.kaggle.com/berkerisen/wind-turbine-scada-dataset}{Kaggle Turkey} \cite{turkeywf} &  1 & Turkey & 40.585469158, 28.990284697 \cite{turkeywf} & 01.01.2018-13.12.2018 & 10min & Active power, wind speed, theoretical power and wind direction & Unknown\\   

\href{https://gitlab.windenergy.dtu.dk/fair-data/winddata-revamp/winddata-documentation/-/blob/master/nm92.md}{NM92} \cite{nm92} &1&Denmark&57.039, 10.075722&19.11.2005-20.01.2006& 10min, raw time series &   Mast, turbine power and load measurements (25 Hz) together with 10-minute statistics of these measurements. Documentation is available \cite{nm92doc}.& DTU Database \cite{DTU}\\ 

\href{https://gitlab.windenergy.dtu.dk/fair-data/winddata-revamp/winddata-documentation/-/blob/master/nordtank.md}{Nordtank} \cite{nordtank}& 1 & Denmark & 55.684436, 12.096689 & 21.10.2004-20.04.2006 &10 min, raw measurements& Mast, turbine power and load measurements (appr. 35 Hz) including 10-minute statistics of these measurements. Documentation is available \cite{nordtankdoc}. & DTU Database \cite{DTU} \\ 

\href{https://ens.dk/en/our-services/statistics-data-key-figures-and-energy-maps/overview-energy-sector}{Denmark Data} \cite{dkmonthly}&$>$5000&Denmark &In \textit{Data on operating and decommissioned wind turbines} location information can be found \cite{dkmonthly}. &2002-2020 &1 month& The dataset is called \textit{Monthly data 2002-2020} and contains yearly production data for all wind turbines $>6kW$. 
&Danish Energy Agency \cite{dkmonthly}\\
\midrule

\href{https://zenodo.org/record/5516550#.YaR-0S9Q2X0}{Wind Spatio-Temporal Dataset2} \cite{ding_yu_2021_5516550}& 200 & Unknown & relative position of the turbines& 01.09.2010-31.08.2011 & 1h & Hourly wind speeds, wind speed and wind direction at three met masts on the same wind farm  \cite{DSWE}& Texas University \cite{ding2019data}\\ 

\href{https://zenodo.org/record/5516543#.YaR-fS9Q2X0}{Wind Spatio-Temporal Dataset1} \cite{ding_yu_2021_5516543}& 120 & Unknown & relative position of the turbines & 01.01.2009-31.12.2010 & 1h & Average wind speed, hourly standard deviation of wind speed \cite{DSWE} & Texas University \cite{ding2019data}\\ 

\href{https://zenodo.org/record/5516541#.YaR-MC9Q2X0}{Offshore Wind Farm} \cite{ding_yu_2021_5516541}& 10 & Offshore, unknown &  relative position of the turbines & 01.07.2007-31.08.2007& 1h & Wind speed, wind direction, air density, humidity, turbulence intensity, above-hub height wind shear, below-hub height wind shear \cite{DSWE} & Texas University \cite{ding2019data} \\ 

\href{https://zenodo.org/record/5516552#.YaSAYC9Q2X0}{Inland Wind Farm Dataset1} \cite{ding_yu_2021_5516552}& 4 + 2 met masts & Offshore, unknown & relative position of the turbines & different time horizons (1yr)& 10min &Wind speed, wind direction, air density, below-hub height wind shear, turbulence intensity \cite{DSWE} & Texas University \cite{ding2019data}\\ 

\href{https://zenodo.org/record/5516554#.YaSAoS9Q2X0}{Inland Wind Farm Dataset2} \cite{ding_yu_2021_5516554}& 4 & Unknown & relative position of the turbines & 2008-2011&  10min  &Wind speed, wind direction, air density, below-hub height wind shear, turbulence intensity \cite{DSWE}& Texas University \cite{ding2019data}\\ 
          
\href{https://zenodo.org/record/5516552#.YaSAYC9Q2X0}{Offshore Wind Farm Dataset1} \cite{ding_yu_2021_5516552}& 2 + 1 met mast & Unknown & relative position of the turbines & 01.01.2009-31.12.2009& 10min &Air density, wind shear, turbulence intensity, humidity & Texas University \cite{ding2019data}\\ 

\href{https://zenodo.org/record/5516554#.YaSAoS9Q2X0}{Offshore Wind Farm Dataset2} \cite{ding_yu_2021_5516554}& 2 & Unknown & relative position of the turbines & 2007-2010& 10min &Air density, wind shear, turbulence intensity, humidity & Texas University \cite{ding2019data} \\

\href{https://zenodo.org/record/5516539#.YaR9zy9Q2X0}{Wind Time Series Dataset} \cite{ding_yu_2021_5516539}& 1 & Unknown & Unknown & 07.10.2014-06.10.2015 & 10min and 1h &Wind speed & Texas University \cite{ding2019data}\\ 

\bottomrule
\end{tabular}
\end{table}

\begin{table}
\footnotesize
\captionsetup{font=footnotesize}
\caption{Overview of the aggregated wind power datasets. The datasets contain spatially aggregated data. A description of the variables can be found in the beginning of this Section. Additionally, the aggregation level is provided in \textbf{Aggregation level}.}
\rowcolors{2}{gray!15}{white}
\label{tab:aggregated}
\begin{tabular}{R{0.11\textwidth}R{0.07\textwidth}R{0.1\textwidth}R{0.1\textwidth}R{0.1\textwidth}R{0.45\textwidth}R{0.25\textwidth}}
\toprule
Dataset &  Location & Aggregation Level& Time span & Interval  & Additional information & Origin \\ \midrule 

\href{https://transparency.entsoe.eu}{ENTSO-E} \cite{entsoe}&European Member States&farm level and other&since 2011&30min to 1h& Wind power data is stored under "Actual Generation Output per Generation Unit". Several software can help to work with the data, e.g. for python entsoe-py can be used \cite{entsoepy}. Data is updated regularly with a lag of approximately five days. A data subset covering the time period from 21.12.2014 to 11.04.2021 is provided by De Felice et al. \cite{entsoezenodo}. A second data subset containing hourly capacity factors for wind onshore from 1982-2019 at national and sub-national (>140 zones) level is  also provided by De Felice et al. \cite{de_felice_matteo_2021_5780185}. Location information is accessible, e.g. by the Joint Research Centre \cite{locentsoe}). & European Union \cite{entsoeorigin} \\ 

\href{https://www.elexon.co.uk/documents/training-guidance/bsc-guidance-notes/bmrs-api-and-data-push-user-guide-2/}{Elexon} \cite{Elexon}&Great Britain&farm level and other&since 2011&30min& B1610 provides wind power generation data. Several software can help to work with the data, e.g.for python the ElexonDataPortal library\cite{elexonpy} can be used. Data is updated regularly with a lag of approximately 5 days. Energy identification codes (EICs) allow for location identification. & Elexon \cite{elexonportal} \\ 

\href{https://www.emi.ea.govt.nz/Wholesale/Datasets/Generation/Generation_MD/}{New Zealand Data} \cite{nzgen}&New Zealand&farm level& since 1997&30min& Generation by plant data is estimated by mapping metered injections into the grid to the generating plant at those injection points. Location information can be found in the network supply points table \cite{nzlocation} and included by mapping POC\_Code \cite{nzgen} and POC code \cite{nzlocation}. The coordinates are then given by NZTM easting and NZTM westing and can be transformed to longitude/latitude using e.g. the python package pyproj\cite{pyproj}. (The current coordinate system is associated with 'epsg:3857', the geodetic world coordinate system with 'epsg:4326'). & Electricity Market Information New Zealand \cite{nzgen} \\ 

\href{http://www.ons.org.br/Paginas/resultados-da-operacao/historico-da-operacao/geracao_energia.aspx}{Brazilian Data} \cite{brgen}&Brazil&federal state &since 2007 &1h to 1 year& \textit{Tipo de Usina} has to be set to \textit{Eólica}.
&Operador Nacional Do Sistema Eletrico (Brazil) \cite{brasiltransparency} \\ 

\href{http://redis.energy.gov.za/electricity-production-details/
}{South African Data} \cite{zagen}&South Africa&province&since 2015 &1h& A manual on how to download the data is also provided \cite{zagendownload}. 
&Department of Energy South Africa \cite{sasource}\\ 


 \href{https://pureportal.strath.ac.uk/en/datasets/australian-electricity-market-operator-aemo-5-minute-wind-power-d}{AEMO} \cite{aus} & South-east Australia &farm level& 01.01.2012-31.12.2013 & 5min & The dataset contains wind power data from 22 wind farms with known location. Abbreviations of the wind farm's names can be found in the csv-file with which locations can be mapped. &AEMO and University of Strathclyde\cite{dowell2015very}\\ 

\href{https://www.dropbox.com/s/pqenrr2mcvl0hk9/GEFCom2014.zip?dl=0}{Gefcom2014} \cite{gefcomdata} \cite{hong2016probabilistic} & Australia & farm level& 01.01.2012-30.09.2012 & 1h & In the dataset wind speeds as u10, u100, v10, v100 are included. The dataset covers 10 wind farms in Australia, their locations are disclosed.& Tao Hong (
University of North Carolina) \cite{hong2016probabilistic}\\ 

 \href{https://zenodo.org/record/4656032#.YZYq8S9Q1hA}{4 Seconds Time Series} \cite{4sec} & Australia & farm level& starting on 01.08.2019 & 4sec & The dataset consists of one wind power time series from a single farm containing 7,397,147 values. &AEMO and Monash University \cite{godahewa2021monash}\\ 
 
\href{ https://zenodo.org/record/4654909#.YZYr_i9Q1hA}{Wind Farm Data with Missing Values} \cite{wfmonash} & Australia & farm level & 01.08.2019-31.07.2020 & 1min & The dataset contains 339 power time series of Australian wind farms with missing values (for some series more than seven consecutive days). Additionally, a second dataset where missing values are replaced by zeros is provided \cite{wfmonashwithout}. &AEMO and Monash University \cite{godahewa2021monash} \\

\href{https://eem20.eu/forecasting-competition/}{EEM2020} \cite{eem2020}  &Sweden& 4 Swedish regions & 01.01.2000-31.12.2001 &  1h& The datasets consists of aggregated data of 4 regions.  &University of Strathclyde  \cite{godahewa2021monash} \\
\bottomrule 
\end{tabular}
\end{table}

\begin{table}
\footnotesize
\captionsetup{font=footnotesize}
\caption{Overview of wind datasets. The separation line divides the global weather datasets from wind measurement datasets and databases. The Orsted dataset is well-known in the wind power research community because the signals are measured close to large offshore wind farms. The last two entries reference databases containing met mast data.}
\rowcolors{2}{gray!15}{white}
\label{tab:wind}   
\begin{tabular}{R{0.11\textwidth}R{0.09\textwidth}R{0.1\textwidth}R{0.1\textwidth}R{0.1\textwidth}R{0.45\textwidth}R{0.2\textwidth}}
\toprule
Dataset / Website & Type & Location & Time span & Interval  &Additional information & Origin \\ \midrule  

\href{https://disc.gsfc.nasa.gov/datasets?project=MERRA}{MERRA} \cite{merradata} & reanalysis & Earth & 1979-02.2016 & 1h to 6h& Several datasets can be accessed, an additional description is provided \cite{merradoc}. MERRA 2 provides data beginning in 1980 and replaces MERRA since 2016 \cite{merra2}. & NASA \cite{merranasa} \\

\href{https://cds.climate.copernicus.eu/cdsapp#!/dataset/reanalysis-era5-land?tab=form}{ERA5} \cite{ERA5} & reanalysis& Earth & 1979 - ongoing& 1h & ERA 5 covers the Earth on a 30km grid from the surface up to a height of 80km \cite{ecmwf_rean}. &ECMWF \cite{ecmwf_rean} \\ 

\href{https://www.ecmwf.int/en/forecasts/datasets/wmo-essential}{ECMWF NWP data} \cite{ecmwf_public}& NWP & Earth & - & 6h and 12h& Two models based on HRES (single high resolution) and ENS (ensemble of forecasts) are publicly available. Both datasets contain mean sea level pressure, geopotential height, temperature and either u and v components of the wind or windspeed on a 0.5° by 0.5°grid. Currently not working (until april: https://www.ecmwf.int/en/forecasts/dataset/wmo-essential)& ECMWF\cite{ecmwf_public} \\ 

\midrule
\href{https://orsted.com/en/our-business/offshore-wind/wind-data}{Orsted} \cite{Orsted} & offshore lidar wind measurements & Baltic sea, Denmark & 2012-2017 & 10min & The three datasets contain data from met masts close to offshore wind farms. They cover different time spans between 2012 and 2017, the largest covers around 2.5 years. Documentation of two of the datasets can be downloaded after registration. Futher information on Fino 2 and the other Fino research platforms in the Baltic Sea is also accessible \cite{fino2}.& Orsted \cite{Orsted} \\

\href{https://energydata.info/dataset}{Energydata.info} \cite{energydata} & met mast measurements & various & - & - & On the page many different wind datasets from different countries all around the globe are provided. Some of them are updated regularly. & World Bank Group \cite{energydataabout}\\ 

\href{http://indecis.eu/data.php}{Tall Tower Set} \cite{talltowers} & high met mast measurements & various & - & - & Tower data of tall towers around the globe. & Supercomputing Center Barcelona \cite{ramon2020tall}\\ \bottomrule 
\end{tabular}
\end{table}

\end{landscape}

\begin{table}
\caption{Overview of synthetic datasets. The datasets contain meteorologically-derived wind power data. A description of the variables can be found in the beginning of this Section. Renewables Ninja can additionally be used to simulate wind farms and create further synthetic data.}
\rowcolors{2}{gray!15}{white}
\label{tab:synthetic}
\begin{tabular}{R{0.1\textwidth}R{0.15\textwidth}R{0.1\textwidth}R{0.08\textwidth}R{0.3\textwidth}R{0.15\textwidth}}
\toprule
Dataset &  Location & Time span & Interval  &Information & Origin \\ \midrule 
\href{https://www.nrel.gov/grid/wind-toolkit.html
}{WIND Toolkit} \cite{datawindtoolkit} & United States (over 126,000 locations) & 2007-2013& 5min & The WIND toolkit is one of the largest synthetic grid integration and wind power datasets publicly available. Several data subsets can be downloaded. The WIND Toolkit is widely used in wind power research (e.g. \cite{howland2019wind} \cite{chen2018model}).& National Renewable Energy Laboratory \cite{draxl2015wind} \\ 

\href{https://www.renewables.ninja/downloads#details-wind}{Renewables Ninja} \cite{ninjadata} & EU-28, Norway, Switzerland & 1980-2016& 1h & Additionally, hourly power output from wind and solar power plants can also be simulated \cite{ren_ninja}. & Iain Staffell and Stefan Pfenninger \cite{staffell2016using} \\ 

\href{https://data.jrc.ec.europa.eu/dataset/jrc-emhires-wind-generation-time-series#dataaccess}{EMHIRES} \cite{emhires} & EU-28, Norway, Switzerland, non EU countries from the Western Balkans & 01.01.1986-31.12.2015& 1h & Data is aggregated by country (onshore and offshore), power market bidding zone and by the European Nomenclature of territorial units for statistics (NUTS) as defined by EUROSTAT \cite{nuts}. A comparison to Renewables Ninja was performed by Moraes et al. \cite{moraes2018comparison} and Gonzalez Aparicio et al. \cite{iratxe2016emhires}. & Gonzalez Aparicio et al. \cite{iratxe2016emhires}\\ \bottomrule
\end{tabular}
\end{table}
However, there is no guarantee for completeness of the tasks discussed here. In the subsequent part, we discuss the spatial and temporal resolution of the datasets, the informative value of location information and the transferability of the results. \\
The spatial and temporal resolution of the individual datasets play an important role in the usefulness and expressiveness of the data. SCADA data is the best with respect to spatial and temporal granularity, it consists of measurements of individual wind turbines and usually preserves a temporal resolution of 10 minutes. The non-SCADA turbine-level datasets have the same spatial resolution but often a coarser temporal resolution. While the spatial resolution within a limited region is high for the turbine level datasets, their spatial coverage is currently low due to data regulations. Therefore, if one aims to predict or analyze the wind power of more than one wind farm, other datasets have to be considered. Among these are the aggregated datasets presented in \Cref{tab:aggregated} which cover large spatial areas. Additionally to spatial coverage, temporal coverage or the time horizon that a dataset covers can also be crucial. For example, seasonal patterns can usually only be detected in datasets that cover every season at least twice, i.e. have a length of at least two years. 
\begin{figure}
    \centering
    \includegraphics[]{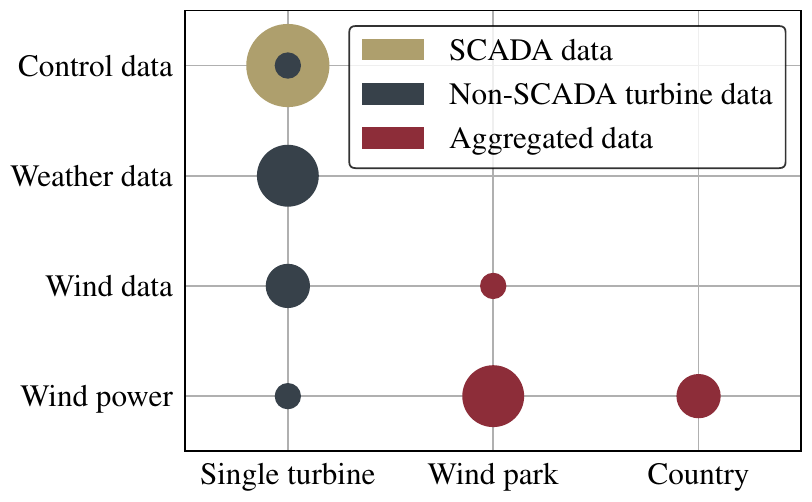}
    \caption{Overview of the spatial aggregation and information contained in the different datasets. The diameter of each circle scales linearly with the number of datasets associated with the respective category. The y-axis is ordered with increasing information content, e.g. datasets that contain control data always contain wind power, wind and weather data as well. While none of the aggregated datasets contains weather data, six of them contain location information that allows to add weather variables.}
    \label{contentbytype}
\end{figure}

As for most countries, wind turbine locations are accessible (see, e.g. the global power plant database \cite{powerplantdb}) the aggregated datasets can be combined with these wind farm locations which can give a broader insight into the wind power network. For three of the datasets (see ENTSO-E, Elexon and New Zealand Data in \Cref{tab:aggregated}), location mapping is straightforward, and we include resources on how to do so. However, this mapping can usually not be performed for most turbine-level non-SCADA datasets found as they lack location information. Furthermore, incorporating control specific parameters is generally not possible afterwards. Nevertheless, some forecasting models explicitly take variables such as nacelle orientation, yaw error or blade pitch angle into account (e.g. Lin et al. \cite{lin2020wind}), and can therefore only be explored, validated and tested using turbine-level data that include these variables.  
In general, the more finely resolved datasets contain more variables and \Cref{contentbytype} shows that, on average, the SCADA datasets contain the most variables while weather data (except for wind data) is never included in the aggregated datasets. 

Another crucial point of wind power forecasting models is their ability to be generalized and transferred to other data. In general, it is hard to elaborate how expressive an individual forecast is. Turbine-turbine interactions and environmental conditions give each turbine and wind farm a unique set of complex physical properties that make evaluating the expressiveness of one turbine's forecast difficult. However, SCADA data for many turbines in larger areas are currently not publicly available, to the best of our knowledge. Therefore, the informative value of SCADA data mostly remains limited to a specific range of conditions. In contrast, aggregated datasets cover larger regions, and when forecasting wind power generation of multiple turbines and parks together, local wind effects can become less relevant. Additionally, Holttinen et al. \cite{holttinen2006prediction} show that forecast errors drop when predicting aggregated wind power production instead of individual sites. Nevertheless, training models on one single wind power dataset always limits their transferability.

To summarize, the application area of the different wind power datasets is mainly defined by the spatial and temporal aggregation level and coverage. Additionally, non-disclosed locations of the turbines play a crucial role as they limit the possibility of including additional weather or terrain data that are often used in more advanced models. However, other models that do not take this information into account are not affected by missing location data. Nevertheless, some variables such as control variables can not be included afterwards and make some datasets unsuitable for testing and validating specific models, independent of the data group to which the datasets belong. Furthermore, the possibility of transferring wind power forecasting models validated on one dataset to another is generally restricted, especially in the case of single turbines and wind farms where local turbulences play a significant role and highly influence the results \cite{marti2006evaluation}. However, taking several sites into account can reduce this effect \cite{holttinen2006prediction}.

\subsection{Wind-based Data}
\begin{figure}
    \centering
    \includegraphics[width=\textwidth]{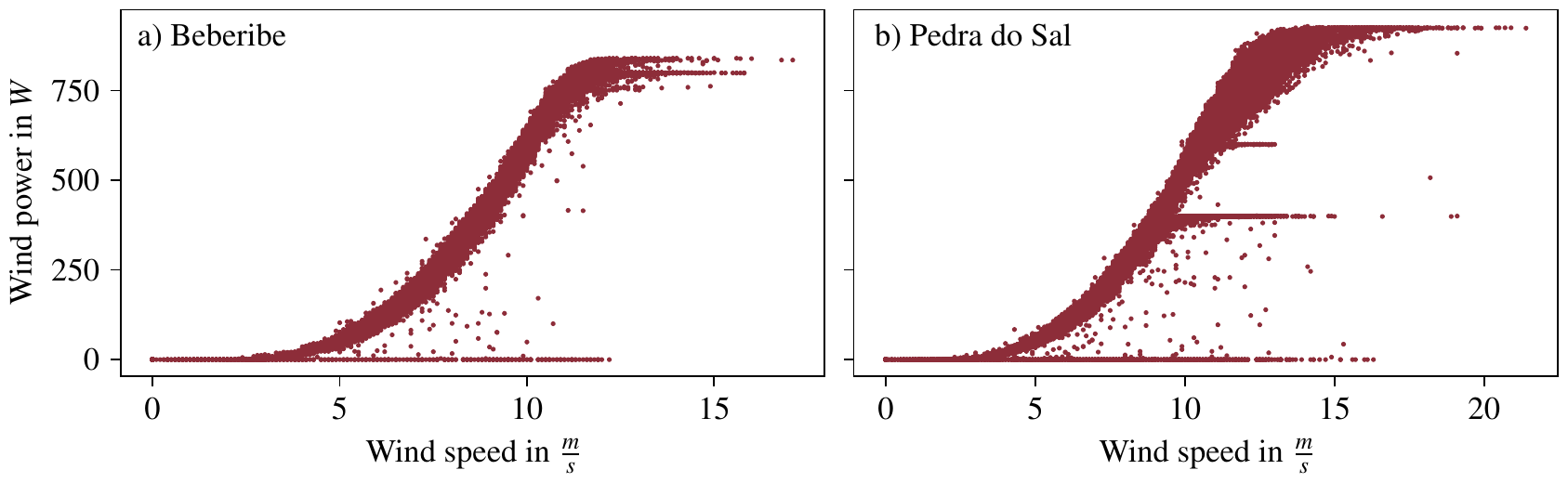}
    \caption{Scatterplot that shows wind power on the y-axis and wind speed on the axis. The plots show data from the first Turbine of the dataset in Beberibe and Pedra do Sal \cite{Beberibe_PDS} respectively. It shows that the mapping from wind speed to wind power is not deterministic.}
    \label{fig:speedpower}
\end{figure}

Especially for larger regions, fine-grained wind power data are not accessible. Whenever this is the case, we can exploit the underlying physical properties of wind power generation. For example, as generated wind power is proportional to wind speed cubed \cite{grogg2005harvesting}, wind speed mainly determines the performance of an operating wind turbine. However, the main advantage of using wind data instead of wind power data is their high spatial coverage. Therefore, some research focuses on wind speed forecasting in order to perform wind power forecasting -- sometimes this is called indirect wind power forecasting \cite{zjavka2018direct}. One example of this is presented by Demolli et al. \cite{demolli2019wind} who aim to predict the wind speed and map it with a turbine-specific power curve to its corresponding wind power. Using such a turbine-specific power curve assumes that this wind speed to wind power mapping is deterministic, i.e. given a wind speed and a power curve the corresponding wind power generation can be computed. However, this assumption usually does not hold (see \Cref{fig:speedpower}). Therefore, other approaches model the wind power curve empirically based on wind data and its corresponding known wind power output to account for this stochasticity. Still, the choice of a power curve model remains difficult; Wang et al. \cite{wang2019approaches} study several power curve models and show that none of the models can outperform all of the other models.

While the expressiveness of wind speed forecasting without additional power data is usually limited, we can perform wind assessment for turbine site selection and efficient planing of wind turbine and wind farm constructions. This is usually done to assess whether the wind speed at a possible future site is high enough to generate a rewarding amount of energy, optimize the turbine locations given the area and location of a planned wind farm, or assess the maximum wind power capacity of a region. For completeness, it has to be mentioned that the maximum and average predicted capacity are not the only relevant considerations for site selection. Additionally to meteorological components, environmental, economic, and societal factors can further restrict site selection \cite{rediske2021wind}.

Instead of using the wind data directly, we can also simulate wind power data based on wind measurements. This simulation allows incorporating the impact of orography into synthetic datasets and can reduce biases \cite{staffell2016using}. Overall, wind data thus has the advantage of covering the whole world, allowing us to use it for indirect wind power forecasting, site selection and synthetic data generation.

\section{Systematic Errors in Wind Data}
\label{sec:syserrors}

There are mainly two problems with the different types of wind data, namely data availability and bias in the data. The first, data availability, mainly affects wind power data (see \Cref{fig:datagroups}) as pure wind data is open-source in many cases. The reason for this is that datasets which include wind power and wind measurements on turbine-level usually belong to wind power companies that do not want to disclose sensitive information about their operations. Without a shift in industry mindset, this lack of available data makes the collection of these datasets in  \Cref{windpower} even more important.
In contrast, the second problem, bias in the data, affects all types of wind data and can be, at least in parts, addressed in the wind power forecasting models themselves. The different wind data measurements include a range of known systematic measurement and modelling errors. As a result, each form of wind data comes with its specific advantages and shortcomings. In the following, we will address these known systemic errors for each form of wind data, namely wind measurements from met masts and turbines as well as computed wind data from NWPs. In order to evaluate the different data types, we first compare wind measurements on turbine-level to wind measurements of met masts and proceed with a discussion of NWP models and reanalysis data.

\subsection{Wind Measurements}
Wind measurements for wind power forecasting are usually performed close to the turbine's nacelle or by close-by independent met masts. The main advantage of nacelle anemometer data is that the wind is measured close to where wind power is generated. However, the moving turbine blades influence these measurements and these measurements are therefore considered inaccurate \cite{antoniou1997nacelle}. Still, turbine-level data is often used to forecast wind power and more recent studies by Cutler et al. \cite{cutler2012using} show an improvement in power curve modelling when using turbine-level wind speed as compared to met mast measurements.
Instead of dealing with these known perturbations induced by the turbine, forecasting with other types of wind data, for example, in the form of close-by met mast LiDAR measurements, can be performed. Some renewable energy companies also support and motivate this approach  \cite{Orsted}. Nevertheless, these wind speeds measured by close-by met masts may not be representative either, especially if the terrain conditions are complex or the met masts are too far away from the turbine of interest \cite{cutler2012using}.
Another shortcoming of many of these met mast measurements is the installation height of the average met mast. With a standard height of 10m above the surface, they do not cover the average turbine height of 50-150m \cite{ramon2020tall}. However, Draxl et al. \cite{draxl2014evaluating} argue that validation against 10 m wind speeds is not sufficient. To address this height-issue Ramon et al. \cite{ramon2020tall} present existing non-standard meteorological observations from 222 taller met masts. Nevertheless, the closeness to turbines and associated wind power generation are usually unknown, their usefulness therefore remains limited.

To summarize, wind measurements can either be taken close to the turbine, where the turbine influences the measurements, or further away from the turbine, where they might not preserve the local wind structure. However, knowing about the challenges these different types of wind data pose allows us to consider them during the development and validation of a forecasting model.

\subsection{Weather Models}
As the number of met masts and monitoring stations is finite, there are no met mast wind measurements for every site. Therefore, globally the density of measurements differs, and weather models covering the whole earth interpolate and model wind and weather data to fill gaps in the measurements. This section will discuss the uncertainties of these weather models, their expressiveness, and some of their known biases.
Commonly used weather models in wind power forecasting are numerical weather prediction (NWP) models \cite{foley2012current}\cite{bossavy2013forecasting} \cite{chen2013wind}. These meteorological forecasting models describe the atmospheric processes with input from weather observations and atmospheric and oceanic simulations. However, due to their nonlinearity, complexity and the gaps in measurements, they are solved with numerical approximations \cite{AlYahyai.2010}. These numerical approximations and the interpolation of data points introduce uncertainty into the NWPs wind predictions. In fact, the NWPs accuracy is generally considered to be the most influential factor on the accuracy and uncertainty of the wind power forecast \cite{sanchez2006short}. However, Holttinen et al. \cite{holttinen2006prediction} show that the errors of the meteorological forecast model drop when data of many sites are being taken into account. Weather models are, in contrast to wind power measurements, available in large amounts covering all countries (see Section \ref{windbased}). However, the underlying meteorologic measurements are of different quality for different regions and can not account for turbine-specific turbulences. Furthermore, they usually contain interpolated data and measurements below hub-height. Therefore reanaylsis or NWP datasets induce biases when they are used to simulate wind power data \cite{staffell2016using}. Reducing these biases is the goal of many forecasting algorithms that rely on weather models \cite{costoya2020using}. Furthermore planetary boundary layer parametrization aims to reduce these errors \cite{draxl2014evaluating}.

Given the two different wind data types, we can decide which data to prefer by comparing the wind power curves that are associated with either wind measurement or modelled wind speed. The more scattered these curves map wind speed to wind power, the harder it gets to forecast wind power given the current wind speed. The wind power curve of the wind speed that is empirically determined using NWP data is more scattered than the power curve that describes the relationship between wind power output and wind speed measured at a reference wind mast \cite{buhan2016wind}. This difference in variance indicates that measured wind speed is more accurate compared to wind speed data from weather models.
Nevertheless, data availability crucially limits the use of wind measurements and therefore none of the wind data types can and should be preferred for each and every application and research question. When working with wind data or simulated wind power data it should be kept in mind that the mapping from wind speed to wind power is non-linear and stochastic. Wind speed forecasting is therefore just a sub-problem of wind power forecasting, not an equivalent.

\section{Discussion}
\label{discussion}
Having presented the open-source datasets and the shortcomings and advantages of the different measurements, we now discuss their use, address their quality and explain how we selected the different datasets. Open-source data and code play an important role in research. While many researchers would like to publish the data they use - we conclude this after various discussions with colleagues - most of the currently used wind power data is subject to confidentiality agreements and therefore non-accessible for the public.
Our main motivation behind this work is therefore to compile a large collection of open-source wind power datasets that helps researchers in the field to use a suitable, non-confidential dataset. Figure \ref{fig:descriptive} shows that the important characteristics such as known location, time horizons of at least one year and fine-grained 10 minute data are covered by several datasets. Nevertheless, the collection lacks a fine-grained ($\leq 10 $min) dataset with large spatial coverage (more than wind farm size). 
We do not discuss the quality or completeness of the data. However, we want to address some points of interest when using the datasets. The definition of data quality depends on the intended task and purpose, and Bessa et al. \cite{bessa2011good} demonstrate that the same holds for forecast qualities. Consequently, incomplete and noisy data usually represent the data measured in the real world better than consistent and complete datasets, although the latter would usually be attributed with a higher quality. The datasets messiness evokes a separate research sub-field, data pre-processing, and various pre-processing techniques are used for example in order to eliminate outliers \cite{morrison2021anomaly} or classify different weather types \cite{he2022combined}. 
Nevertheless, the credibility of the data depends on the data's origin and should also be considered as a part of data quality and reliability. Therefore, it has to be mentioned that the Kaggle datasets \cite{Kaggle_FD}, \cite{turkeywf}, \cite{wpf2} except for the dataset of GEFCom2012 \cite{gefcomdata}, lack a good documentation of the variables and the datasets' origins are unknown. \\
We compiled the datasets listed in this paper in several different ways: By searching online for datasets, getting in contact with wind power forecasting researchers from every continent to ask for available open-source data and energy data regulations in their region, and searching for papers that work with disclosed data. There is no guarantee for the completeness of the data. While we did not find other SCADA datasets, more countries probably provide or will provide some form of aggregated wind power data, and we know that many more resources of wind data exist. While we try to present all datasets that are hard to find and of high interest, such as the turbine-level datasets in Section \ref{windpower}, the aggregated and wind-based datasets should be seen as representatives of their corresponding groups. In these categories, we tried to cover different spatial and temporal resolutions. 

\section{Conclusion}
\label{conclusion}
Wind power forecasting is an important tool that can help to integrate wind power efficiently into the power grid. In addition, wind power forecasting can be used to make turbine-specific adjustments and the mapping of wind speed to wind power allows for wind power resource assessment. All of this makes high-quality wind power forecasting and other forms of renewable energy forecasting important in the development of a more sustainable power grid. In this paper we provide a categorization and overview of open-source wind power datasets that can be used for various wind power forecasting tasks. Among these are wind power forecasting on different time and aggregation scales, wind ramp forecasting, wind variability forecasting and wind turbine condition monitoring.
Wind power forecasting is a hard task and its accuracy and quality does not only depend on the aim of the forecast and the quality of the data but also on the real-world properties that the data aims to represent. Among these are for example the climatic characteristics of the country or region that is studied, terrain complexity and atmospheric (in-)stability and site specific turbulences. This limits the comparability of models that were tested and validated on different datasets -- the transferability of results is therefore hard to assess. A first step towards facing this issue can be taken by using open-source data and disclosing the research process. 
In order to make this search for datasets easier, we present the, to our best knowledge, largest variety of different datasets that can be used for different wind power forecasting tasks. Our categorization into five different groups of data such as fine-grained turbine-level power data, aggregated datasets or even synthetic data allows to find a suitable dataset for various different tasks. With overall more than forty open-source datasets we claim that for many tasks in wind power forecasting suitable datasets do not have to be confidential. Future work will benchmark different forecasting models on the open-source wind power datasets in this overview.

\section*{Acknowledgements}
Funded by the Deutsche Forschungsgemeinschaft (DFG, German Research Foundation) under Germany’s Excellence Strategy – EXC number 2064/1 – Project number 390727645 and the Athene Grant of the University of Tübingen. The authors thank the International Max Planck Research School for Intelligent Systems (IMPRS-IS) for supporting Nina Effenberger.

\bibliographystyle{abbrv}
\bibliography{main}

\end{document}